\documentclass{IEEEtran}

\usepackage[utf8]{inputenc} % allow utf-8 input
\usepackage[T1]{fontenc}    % use 8-bit T1 fonts
\usepackage{hyperref}       % hyperlinks
\usepackage{url}            % simple URL typesetting
\usepackage{booktabs}       % professional-quality tables
\usepackage{amsfonts}       % blackboard math symbols
\usepackage{nicefrac}       % compact symbols for 1/2, etc.
\usepackage{microtype}      % microtypography
\usepackage{subfiles}
\usepackage[export]{adjustbox}
\usepackage{graphicx, subcaption}
\usepackage{amsmath}
\usepackage{wrapfig}
\usepackage{multirow}
\usepackage{floatrow}
\usepackage{caption}
\usepackage{booktabs}
\usepackage{subcaption}

\title{Multi-Density Sketch-to-Image Translation Network}

\author{%
    Jialu Huang, Jing Liao, Zhifeng Tan, Sam Kwong, Fellow, IEEE\\
}

\begin{document}
\maketitle
\begin{figure*}
    \centering
    \includegraphics[width=0.8\textwidth]{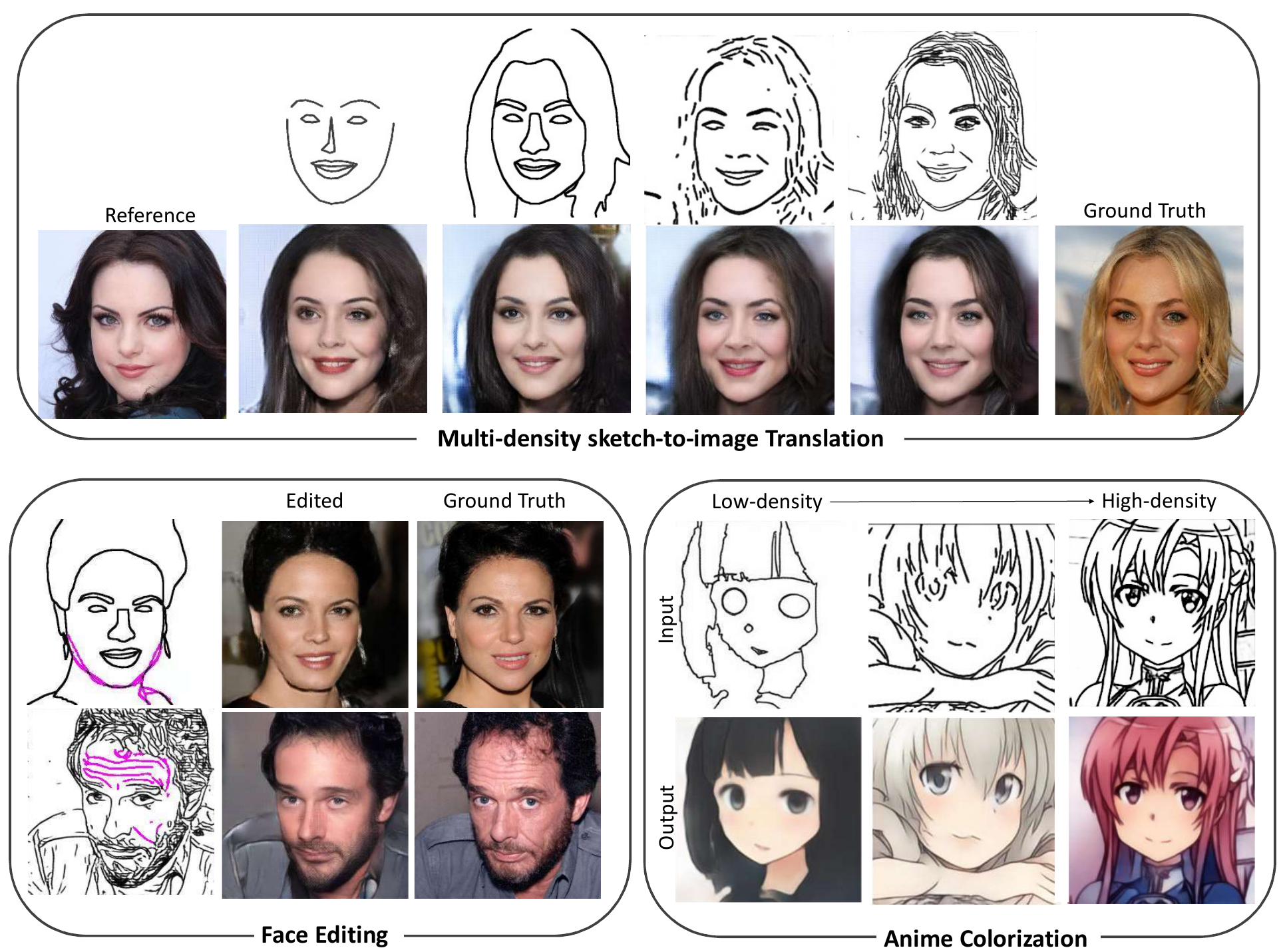}
    \caption{\small{We propose a multi-density sketch-to-image translation network (MDSIT), which can support input sketches at different density levels (Top). Our MDSIT can be applied in applications including face editing and anime colorization, and also provide coarse-to-fine levels of controls to these applications (Bottom).}}
    \label{fig:T}
\end{figure*}
%\twocolumn[{%
%\renewcommand\twocolumn[1][]{#1}%
%\maketitle
%\begin{center}
%    \centering
%    \includegraphics[width=0.9\textwidth]{fig/T.pdf}
%    \captionof{figure}{\small{We propose a multi-density sketch-to-image translation network (MDSIT), which can support input sketches at different density levels (Top). Our MDSIT can be applied in applications including face editing and anime colorization, and also provide coarse-to-fine levels of controls to these applications (Bottom).}}
%\end{center}%
%}]

\begin{abstract}
Sketch-to-image (S2I) translation plays an important role in image synthesis and manipulation tasks, such as photo editing and colorization. Some specific S2I translation including sketch-to-photo and sketch-to-painting can be used as powerful tools in the art design industry. However, previous methods only support S2I translation with a single level of density, which gives less flexibility to users for controlling the input sketches. In this work, we propose the first multi-level density sketch-to-image translation framework, which allows the input sketch to cover a wide range from rough object outlines to micro structures. Moreover, to tackle the problem of noncontinuous representation of multi-level density input sketches, we project the density level into a continuous latent space, which can then be linearly controlled by a parameter. This allows users to conveniently control the densities of input sketches and generation of images. Moreover, our method has been successfully verified on various datasets for different applications including face editing, multi-modal sketch-to-photo translation, and anime colorization, providing coarse-to-fine levels of controls to these applications. 
\end{abstract}

%\keywords{Multi-scale disentangle, GAN, Deep Image Synthesis, Interactive Editing}
\let\thefootnote\relax\footnotetext{This work was supported in part by the National Natural Science Foundation of China Grant 61672443, in part by Hong Kong GRF-RGC General Research Fund under Grant 9042322 (CityU 11200116), Grant 9042489 (CityU 11206317), and Grant 9042816 (CityU 11209819)\par
This work was partly supported by Hong Kong ECS grant No. 21209119,
Hong Kong UGC and Start-up grant No. 7200607, CityU of Hong Kong.\par
Jialu Huang, Jingliao and Sam Kwong are with the Department of Computer Science, City University of Hong Kong, Kowloon, Hong Kong (e-mail: jialhuang8-c@my.cityu.edu.hk, jingliao@cityu.edu.hk, cssamk@cityu.edu.hk) Sam Kwong is also with City University of Hong Kong Shenzhen Research Institute. Zhifeng Tan is with Brion-ASML (e-mail: steven.tan@asml.com)}

\newcommand{\Lim}[1]{\raisebox{0.5ex}{\scalebox{0.8}{$\displaystyle \lim_{#1}\;$}}}
\section{Introduction}
With the boom of deep learning, image-to-image (I2I) translation achieves outstanding results in both terms of quality and diversity, one of its important sub-tasks that S2I translation has been well studied and developed as auxiliary tools in many applications. The pioneering work Pix2Pix \cite{pix2pix} uses a conditional GAN to generate images based on a given input, later, BicycleGAN \cite{bicyclegan} extends this work to a multi-modal framework which can generate images with different styles. Some other state-of-the-art I2I translation methods including MUNIT \cite{munit} and DRIT \cite{drit} generate images based on different references with un-paired data. Although those classic I2I methods can achieve high quality results, when performing the S2I translation tasks, they can only support input sketches in a single density level. However, in practice, different densities of sketches are necessary for representing the coarse-to-fine levels of information to meet the requirements of different applications and users. For example, in the art design process, sketches are in nature created from the coarse level to the fine level. It would be simple outlines of objects at the beginning, and detailed information like textures and micro structures would be added later. Some middle results of those sketches might be applied to S2I translation to obtain some preliminary design output rather than performing the translation when the sketch is complete. Moreover, artists have various styles that some of them prefer more abstract sketches with low level of density while others tend to add more details into the sketches. Therefore, previous methods only support sketches with identical density level can no longer meet the requirement. Taking multi-density sketches as input is challenging for two major reasons. First, it is difficult to collect multi-density sketches corresponding to one image in the training stage, and only with a finite number of training sketches it still cannot support input at continuous density level. Second, balance between the style and content information in the generated results are needed. Specifically, with a coarse level sketch, more information are obtained from the reference, compared to a fine level sketch itself contains style information like textures, light condition, etc.\par

In this work, we propose the first Multi-Density Sketch-to-Image Translation (MDSIT) framework. Within a single model of MDSIT, the input sketches can be selected in a wide range from rough object outlines to sketches with full details. Moreover, we project the density into a continuous latent space, which can then be linearly controlled by a parameter. This allows users to conveniently control the image generation process. Our MDSIT consists of two sub-networks: a Multi-Density Sketch Generator (MDSG) which maps parameters to the continuous sketch representation space; and a Multi-Density Translation Network (MDTN) that could generate images conditioned on different levels of sketches outputted by MDSG.\par

The MDSG is proposed to tackle the problem of lacking sketches with continuous densities in the training stage. We first construct a set of sketches with different density levels via a traditional sketch extraction method that coherent line drawing (CLD), and then we design the MDSG to encode those images into a linear space where sketches with continuous densities can be decoded. Since continuous ground-truth sketches are not available in the training stage, the MDSG is semi-supervised and it is important to provide a proper constraint when there is no ground truth. Inspired by the InfoGAN \cite{infogan}, our MDSG takes a scalar together with the given image as the input, and the output sketches will then be encoded again into the input scalar. With this scalar reconstruction, the network can learn the distinctions among different levels of sketches. To improve the linearity of the MDSG, we also add an adaptive feature distance objective function that tries to minimize the embedding distance in the linear space. Although previous methods like canny edge detection and CLD may be able to generate continuous sketches, they suffer from serious performance issues which cannot be applied in online training. Moreover, it may take time to find proper parameters in such methods to generate continuous sketches. \par

\begin{figure*}[ht]
    \centering
    \includegraphics[width=\textwidth]{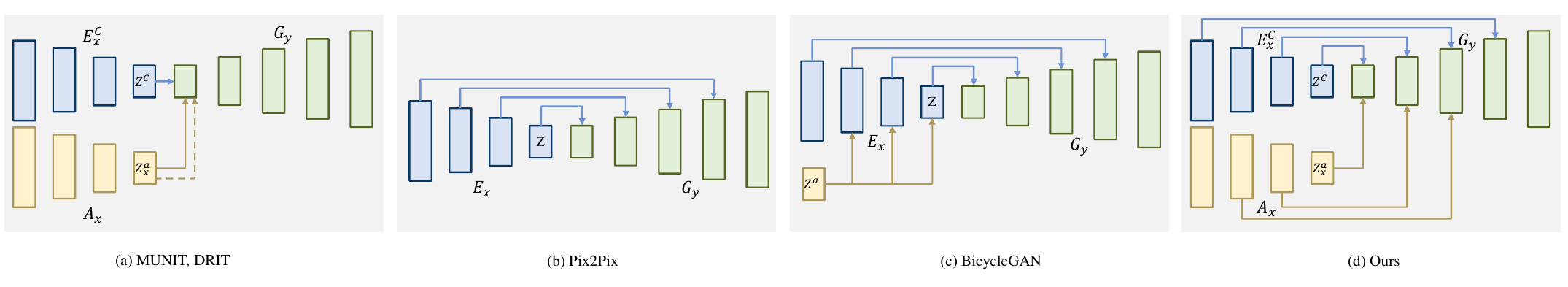}
    \caption{\small{Different ways of combining latent features for generation in I2I translation network, where $E_x$ is the encoder, specifically,  $A_x$ is the appearance encoder in $X$ domain, and $G_y$ is the generator/decoder that can generate images in $Y$ domain. $Z$ stands for the latent features. In models like MUNIT\cite{munit} and DRIT/DRIT++\cite{drit, drit++} with the assumption that images can be encoded into content and appearance latent space, $E^c$ stands for the content encoder, $Z^c$ and $Z^a$ stand for content and appearance code, respectively. In addition, solid lines indicates direct combination like concatenation while dot line represents adopting the style code via AdaIn \cite{adain} layers, both methods can achieve the injection of single-level style code.}}
    \label{fig:style_comparison}
\end{figure*}

However, even with the training sketches, the multi-density input brings extra challenges to the S2I translation that different densities of sketches should correspond to different distributions of style code which is difficult to be achieved by the bottle-neck style feature. Recent efforts have been made to address the issue of conditional image generation shown in Fig. \ref{fig:style_comparison}, a general way is to inject the extracted style code into the decoder, which can be done via concatenation \cite{pix2pix} or using AdaIn \cite{adain} layers (dot lines) \cite{munit}. Moreover, some methods like pix2pix \cite{pix2pix} use Unet \cite{unet} structure to ensure the generated images conditioned on the given input. We adopt this structure in the MDTN as it aims to keep the information in the given sketch as much as possible. Different from these methods that only using a single style code, we inject the style code from different convolution layers into the decoder, and the content represented by multi-level feature maps will merge the corresponding style information, this can help balance the style information based on different levels of sketches.   \par
 
We demonstrate the effectiveness of our method in several datasets, besides real faces, we apply our method to the anime and cityscape (GT) for colorization and editing, which demonstrate the generalization capability of our method. The major benefit of our MDSIT in these applications shown in Fig. \ref{fig:T} is to provide coarse-to-fine levels of control.\par

To summarize, our main contributions are concluded as follows:
\begin{itemize}
   \item We propose a method to generate multi-density sketches and allow continuous control with parameters.
    \item We propose the first framework to support multi-density sketch-to-image translation
    \item Our framework is successfully applied to many applications including face editing, colorization and editing, and provides coarse-to-fine levels of controls to these applications.
\end{itemize}

\section{Related Work} \label{sec:rela}
\textbf{Image-to-image translation.} I2I translation has been a heat topic recently that can be applied to lots of interesting applications such as super-resolution \cite{mmapp_sr1, mmapp_sr2}, de-rain \cite{mmapp_derain, mmapp_derain2}, conditional image synthesis \cite{mmapp_anime, mmapp_face, mmapp_drivescene, tmmapp_cloth, tmmapp_face}, etc. Most I2I translation models \cite{pix2pix, cyclegan, bicyclegan, unit, munit, drit, tmm_I2I1, tmm_I2I2, tmm_I2I3, tmm_segin} are designed from an autoencoder structure where images are first encoded into a bottleneck features and then decoded to images in another domain. According to the ways of combining latent features for generation, these networks \cite{munit, drit, drit++, pix2pix, pix2pixhd, bicyclegan} can be categorized into different types, as shown in Fig. \ref{fig:style_comparison}. MUNIT \cite{munit}, DRIT/DRIT++ \cite{drit, drit++} encode an image into two latent spaces namely content and style space, the two bottleneck features are then combined and decoded into images in another domain. Since they only use the bottleneck features for decoding, we classify this type of models as the single-level style and single-level content model. Although the generated images are visually similar to the input images and the references, it cannot preserve fine details in both content and style due to data compression in the encoding process. Pix2pix \cite{pix2pix} encodes images into one latent space, and in the decoding stage, it uses the Unet structure that injecting features from the encoder at different levels into the decoder to generate final results. This Unet structure ensures different levels of information in the input image can be inherited by the generated result and it can be classified as the multi-level content generation model. However, pix2pix is a single modal network which cannot be guided by a given style reference. To address this limitation,  BicycleGAN \cite{bicyclegan} adopted a single-level style and multi-level content framework. Although it proposes to inject the style code into different levels of encoding layers, the injected style codes at different layers contain the same information. In summary, either single-level content or single-level style (or no style) is used in existing I2I translation networks, thus they cannot achieve high quality when applied to the multi-density S2I translation task which requires combining different levels of content and style information for different density inputs. Therefore, we proposed a novel multi-level content and multi-level style network MDSIT as shown in Fig.\ref{fig:style_comparison} (d), which incorporates different levels of features in both content and style aspects to meet the multi-density requirement.   \par

 \textbf{Sketch extraction.} Our method takes multi-level sketches as the input for the multi-density S2I translation, and thus it requires to extract the sketch from images for the training purpose. One option of generating sketch data is to modify and refine edge detection results. There are lots of classic edge detection methods proposed in the past decades including Canny's \cite{canny}, mean-shift segmentation \cite{meanshift}, DoG \cite{dog}, etc. However, those methods would produce short isolated edge fragments, especially in the areas with image noise or low contrast. In addition, those conventional methods could not capture semantically and perceptually interesting features as the edge maps are generated based on thresholds. \cite{cld} proposed a new method that Coherent Line Drawing (CLD) presenting the flow-driven anisotropic filtering framework, which can generates high-quality line drawing with a group of smooth, coherent, stylistic and natural lines. Furthermore, \cite{hed} proposed an edge detection algorithm based on deep convolutional neural network (DCNN). HED demonstrates that different feature layers can provide different scales of information, therefore, the side-output of HED are a set of multi-scale sketches. However the scale of HED is discontinuous. So we adopt both CLD for generating some discrete scales of sketches and then use our MSCG to continuously interpolate between them.

\section{Method}
\begin{figure*}[ht]
    \centering
    \includegraphics[width=\textwidth]{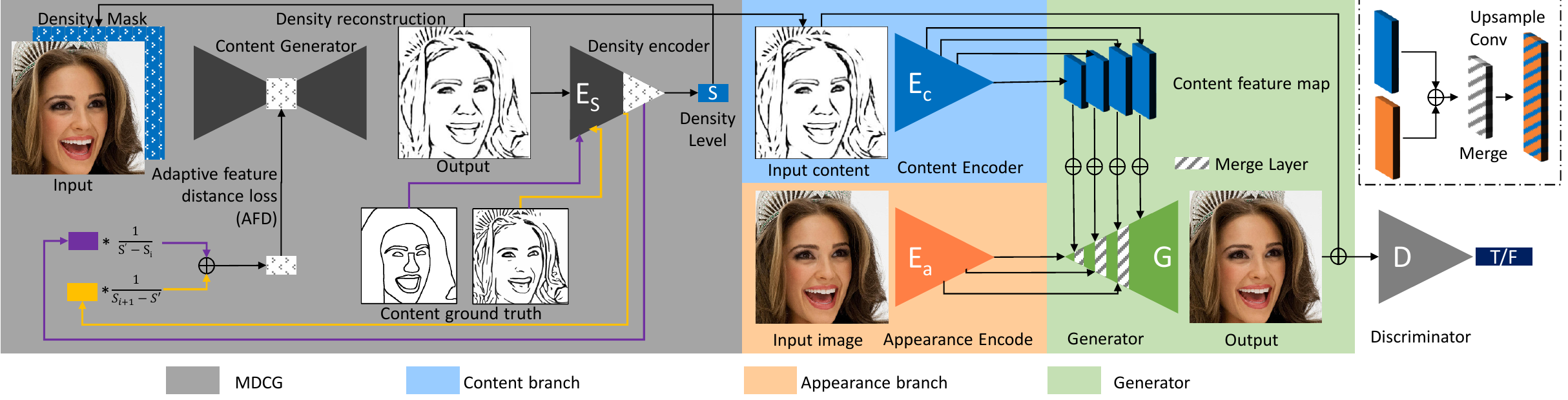}
    \caption{\small{MDSIT consists of two sub-network, namely the MDSG (left) and the MDTN (right). First, we sent an image with a density factor in the form of a scale mask to the MDSG where it can generate a content instance at the given scale. The density factor here can control the level of complexity of the content instance. Later, the MDTN recombine input images with the guidance of continuous sketch instances generated by the MDSG.}}
    \label{fig:framework}
\end{figure*}

Our Multi-Density S2I Translation Network should be trained with multi-density sketches. To tackle the problem of lacking training data, we first employee traditional edge detection methods to generate several key density sketches, and then design the MDSG to learn the in-between sketches. To support the later online training of the translation network, we map a parameter to the density hence we have more intuitive and flexible control to the sketch generation by indicating specific density scale. With the training data, we then design the MDTN network for the translation task. We start with multi-modal I2I translation network like MUNIT\cite{munit} and DRIT/DRIT++\cite{drit,drit++}, which only use a single layer of style and content feature for decoders. However, we found a single layer of the style code might not contain enough information for the case that a rough sketch is used as input. In addition, some low level features of input sketches are missed in the generated results when only using the bottle neck content code. Therefore we propose combining multiple style and content features in decoder,  and the generated results can maintain as many as possible details of the input sketch and meanwhile the more style information can be provided when the input sketch is sparse.
To summarize, we propose a multi-density translation network MDTN and a linear sketch image generator MDSG, we then fine-tuning the MDTN based on the output of MDSG to make sure the output image can be generated conditioned on any density of sketches. The framework of the MDSIT is shown in Fig.\ref{fig:framework} \par

\subsection{Multi-Density Translation Network}
First, we define the multi-density S2I translation problem in a formal way. Let $X\in\mathcal{X}$ be an input reference image and $Y\in\mathcal{Y}$ stands for a sketch images where $Y(s)$ represents a sketch with density $s$. Let $y$ and $z$ be the latent representation of the content and appearance of the image, respectively. Given a sketch image $Y(s)$ with the density $s$, the content encoder $E_{c}$ first extracts its content feature set $y(s) = \{y(s)_0, y(s)_1, ..., y(s)_n\}$, where $y(s)_i$ indicates the feature at the $i^{th}$ layer, ($i\in N)$). While in the appearance branch, the input reference image $X$ is sent to the appearance encoder $E_{a}$ to extract necessary style features $z = E_{a}(X)$. Similarly, the appearance latent feature $z$ also contains multi-layer information that $z = \{z_0, z_1, ..., z_n\}$\par

Note that the generator $G$ can be decomposed into a series of intermediate layers including merge layers $M(\cdot)$, residual layers $G_R=\{G_{R0}, G_{R1}, ..., G_{Rn}\}$ and an output layer $G_{output}$. The merge layer shown in Fig.\ref{fig:framework} (right side) is used to merge the appearance information with given content features. For the first activation layer, $m_0 = M_0(y(s)_0, z_0)$. While the later layers take both latent features and the output of the previous layer as input:
\begin{equation}
    m_i = M_{i}(y(s)_i, z_i, m_{i-1})
    \label{eq:zi}
\end{equation}
After that, the last merge layer is sent to residual layers and then the output layer to generate the final image $I = G_{output}(G_{R}(m_n))$. Since the real input in this process are two sets of feature codes $y(s)$ and $z$, we refer this whole process in the following sections as $I = G(y(s), z)$.\par
With the density conditioned content and appearance latent features, we use $G$ to reconstruct the input image $\hat{X} = G(y(s), z)$. Although the style information obtained from the appearance branch are independent to the given density, the merge layer can take associated style information based on the given content, this makes the decoding process smoother. To ensure the results inherits all the characteristics from the given sketches, more skip-connection layers are used compared to the style feature layers.\par  
To reconstruct the image $X$, we apply several losses to supervise the training procedure. We apply the L1 loss to restore the input image, shown in Eq.\ref{eq:l1}.
\begin{equation}
    L_{recon} = \mathbb{E}_{y, z, \mathcal{G}}[||X - G(y(s), z)||_1]
    \label{eq:l1}
\end{equation}
In addition to the L1 loss, we also use a feature loss Eq.\ref{eq:lf} to constraint the reconstructed images in feature space. where $\Phi(.)$ indicates the feature extraction from the VGG19 \cite{vgg}.
\begin{equation}
   L_{\mathcal{F}} = \mathbb{E}_{y, z, \mathcal{G}}[||\Phi(X) - \Phi(\mathcal{G}(y(s), z))||] 
   \label{eq:lf}
\end{equation}
We also employ a multi-scale discriminator $D$ to increase the fidelity of the generated images. Apart from generating nature and reliable images, we also want the discriminator to evaluate if the generated images are conditioned on the given content, therefore we concatenate the content with the generated image as a fake sampled and send it to the generator. The adversarial loss is defined as the follows:
\begin{equation}
    \begin{split}
    \mathcal{L}_{adv}(G, D) &= \mathbb{E}_{x,y}[logD(X, Y(s))]\\ 
    &+ \mathbb{E}_{X, y, z, \mathcal{G}}[log(1-D(\mathcal{G}(y(s), z), Y(s)))] 
    \end{split} 
\end{equation}

\subsection{Multi-Density Sketch Generator}
Our goal for MDSG is to generate sketches with continuous density, which is challenging due to the following reasons: (1) obviously, it is impossible to obtain continuous ground truth images for every scale factor, which means the MDSG is a semi-supervised model; (2)moreover, we aim to generate multi-density sketches and use a scale factor to control its visual complexity, it is tricky to build up a precise connection between a high dimensional distribution (image) and a scalar. (3) in addition to mapping the generated content image to a scale factor, we try to ensure the linearity and the continuity in the image space, that is, to keep the variation in the image space as similar as possible to the variation of the according scales. This can help for estimating the visual complexity in the sketch generation stage, though non-linear mapping between the scalar and sketch densities can also be adopted, it is difficult to estimate the actual density of output with a given number.\par
Although there is no continuous content image, we can still use several sketch images that describe the essential semantic information of the input image as the key density images $K = \{K_1, K_2, ..., K_n\}$, $K\in \mathcal{Y}$, which are corresponding to a set of sampled density $s = {s_1, s_2, ..., s_n}$ . Then the task now is changed to generate semantically continuous content images between two key density images. As shown in Fig. \ref{fig:framework}, we first extend the density factor to a density mask $M_s$ by filling the mask with the density factor, combined with the reference image, and send them into the content generator $G_c$, where $(\hat{Y}(s)) = G_c(M_s, X)$. Considering the situation that we have the ground truth image, and then an image reconstruction loss can be used to reconstruct the key density images: 
\begin{equation}
    L_{recon_c} = \mathbb{E}_{x, s, \mathcal{G}_c}||K_i - G_c(M_{s_i}, X)|| 
    \label{eq:lc}
\end{equation} 
Since there is no such ground truth images between $K_i$ and $K_{i+1}$, it is a semi-supervised problem requiring indirect control. Inspired by InfoGAN \cite{infogan}, we try to build a connection between the generated sketch images and its corresponding density factors by adopting a density encoder $E_s$, which aims to encode the generated sketches $(\hat{Y}(s))$ back to a density scale that $\hat{s} = E_s((\hat{Y}(s)))$. Since we want explicit control over the generated images, we And then it is optimized by a scale reconstruction loss:\par
\begin{equation}
    L_{recon_s} = \mathbb{E}_{x, s, \mathcal{E}_s, \mathcal{G}_c}[||s - E_s(G_c(M_s, X))||]
\end{equation}

where $K_i$ and $Y(s_i)$ indicate the same sketch image and ${\hat{Y}(s_i)}$ is the reconstruction of $Y(s_i)$.\par
In our initial experiments, we found that the scale reconstruction loss can only help to estimate the sketch images around key density sketches. If there are large geometric or semantic variation between two key density images, the density encoder as well as the scale reconstruction loss are no longer able to ensure the linearity of images between the two key density sketches. Therefore, we design an adaptive feature distance loss (AFD) to force the network lean the correlation between two key density sketches.\par
For a scale $s'$ matching the non key density sketch $Y(s')$. It has two neighbor key density sketches $K_i$ and $K_{i+1}$ that are corresponding to two density factors $s_i$ and $s_{i+1}$, where $i = h(s^{(i)})$, we defined the AFD loss as following:
\begin{equation}
    L_{AFD} = ||\Gamma(Y(s')) - (\frac{1}{s' - s_i}\Theta(K_i) + \frac{1}{s_{i+1} - s'}\Theta(K_{i+1}))||
\end{equation}
where $\Gamma(\cdot)$ and $\Theta(\cdot)$ indicate the feature extraction from the encoder of content generator $G_c$ and the scale encoder $E_s$. The AFD loss ensures the linearity in latent space as the adaptive weights are inversely related to the distance between current content image and its neighbor key density sketches. In other words, the generated sketch is forced to be of higher similarity to its closest neighbor key density sketches. Our experiments also show that the AFD loss significantly improves the continuity of generated content images. 

\subsection{Data Construction}
 Although we cannot prepare continuous sketches from simple structure to the fine sketch with detailed lines, we first try to train the network with several sketches with visual complexity differences. To be more specific, we take the data construction in CelebA-HQ\cite{CelebAMask-HQ} as an example, this process is similar to that in other datasets. For the first level sketch, it should contain basic information of human face such as pose, basic structure and expressions, and we found the landmark can meet those criterion. Therefore, for the first level, we interpolate lines between landmark points detected by open source algorithm \cite{dlib}. In addition to the basic information, shape of facial features, hair and accessories should be covered in the next level. We found the facial parsing results can perfectly meet such requirement as the parsing mask can describe coarse shapes of the facial features without specific lines. \par
\begin{figure}[ht]
    \centering
    \includegraphics[width=\columnwidth]{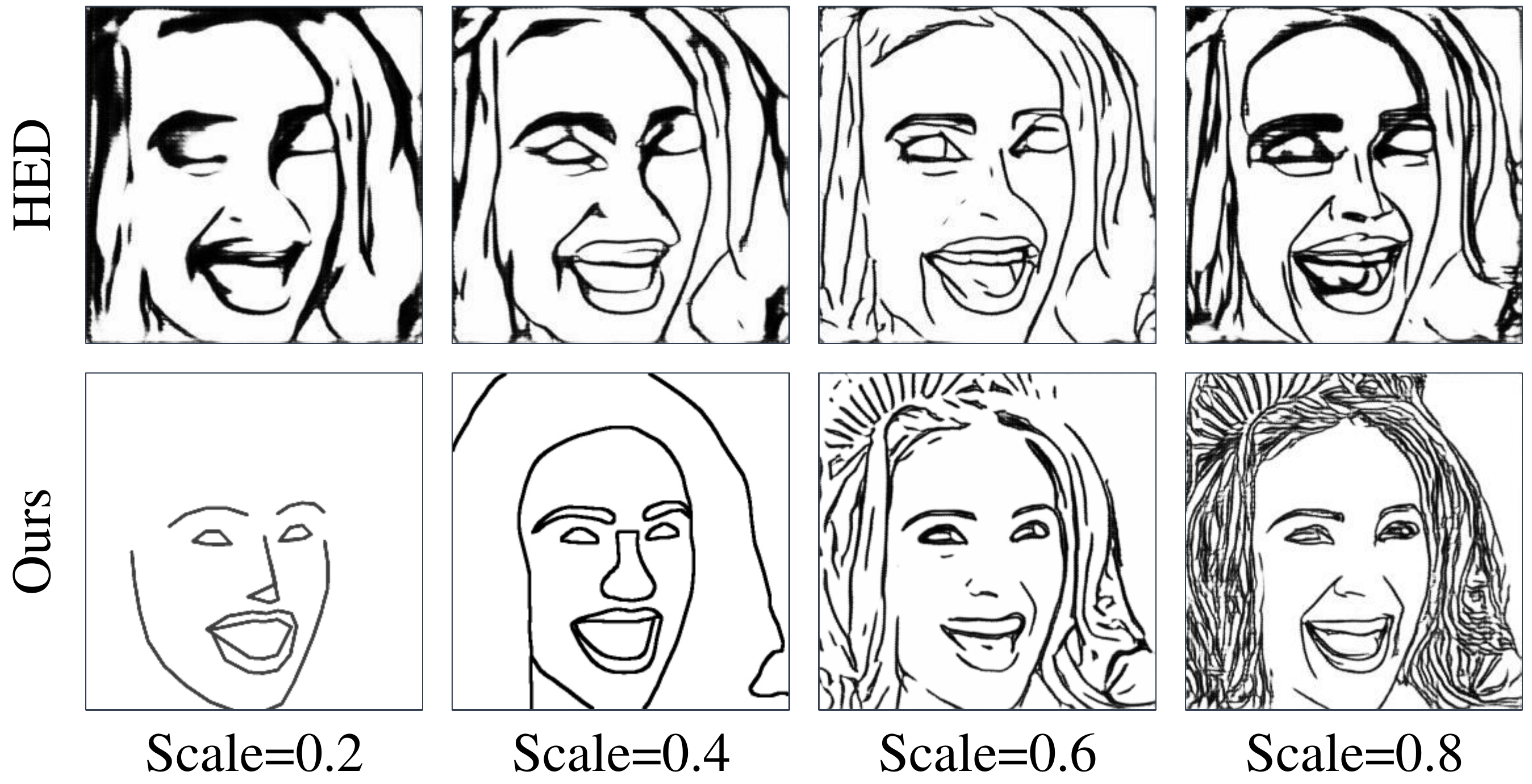}
    \caption{\small{An example of the key density sketches}}
    \label{fig:face_content}
\end{figure}
As for the latter levels, we try to add details including particular hair styles, skin textures (e.g. winkles) and even light conditions into the sketches. We first tried HED \cite{hed} that aims for extracting different levels of edges from general image, however, it is not suitable for our cases as the sketches extracted by the HED contain lots of grey pixels and sketches at different levels are perceptually similar to the neighbours.\par
We then adopt a conventional algorithm that Coherent Line Drawing (CLD) \cite{cld} to generate the level3 and level4 sketches. By adjusting the parameters of CLD algorithms such as ETF kernel size, degree of coherence, noise level and threshold values, we can control the complexity of the generated line drawing. Finally, we use a sketch simplification method proposed by \cite{sketchsimplify} to refine the sketches in all levels, which adjusts sketch thickness, removing unnatural short lines and noise, makes the final sketch shown in Fig.\ref{fig:face_content} smooth and clear.  \par

\subsection{Fine Tuning Step}
In the training stage, the MDTN is trained only with key density sketches. Therefore jointly training is then needed to fine tune the MDTN, so that it can take sketch images at any density as the input and also be controlled by a single scale factor rather than a visual instance. In the tuning stage, we fixed the parameters of MDSG and generate content images with a random scale, which is taken as content input $y(s)$ for MDTN. Without the fine-tuning, the MDTN may not accurately reconstruct the image based on a non key density sketch since it never appears in the stand-alone training procedure.  After the fine-tuning step, the translation can be controlled by a continuous density factor. \par

\section{Discussion}
\subsection{How multi-density translation works in MDTN?}
\begin{figure}[ht]
    \centering
    \includegraphics[width=\columnwidth]{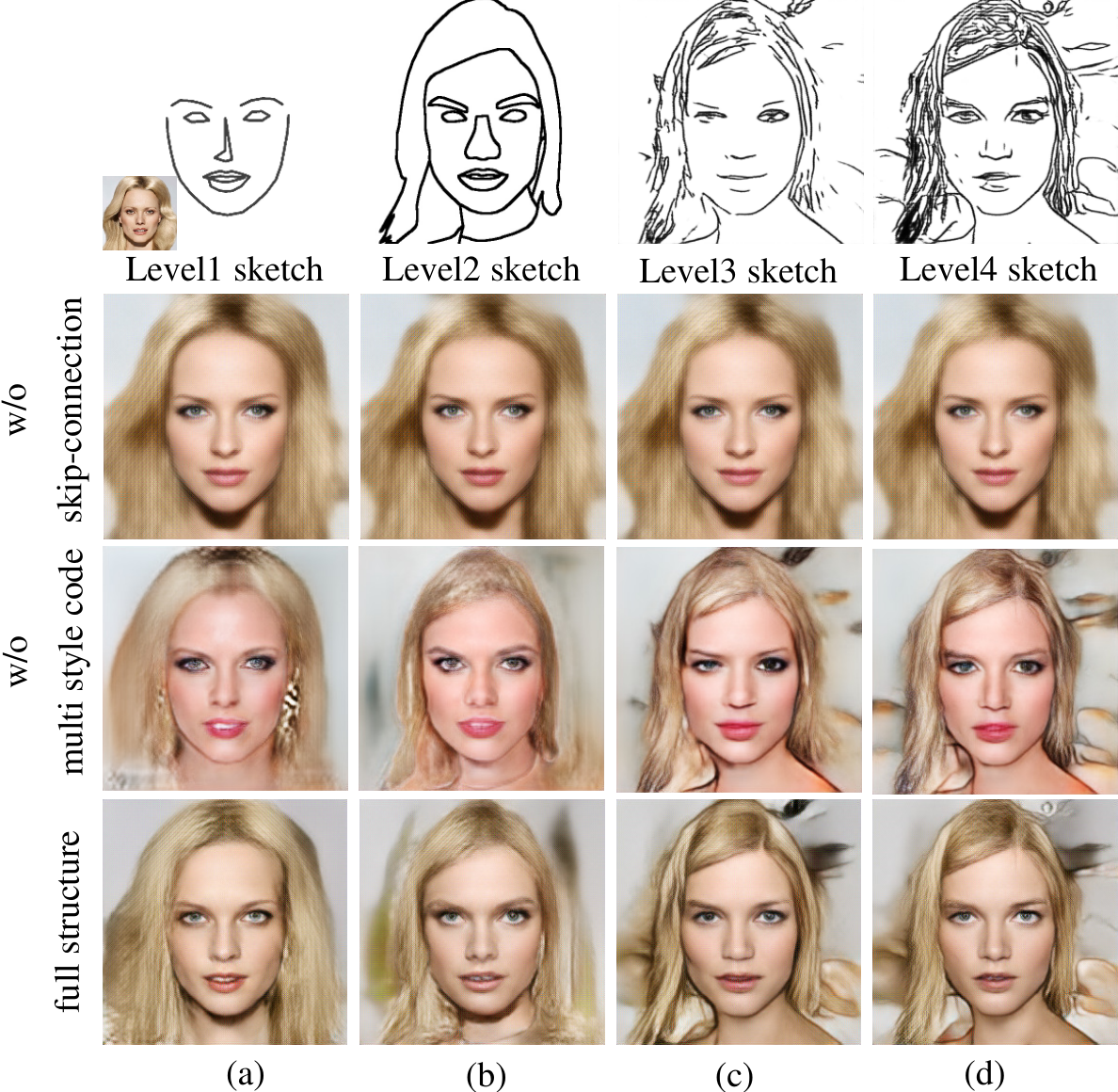}
    \caption{\small{Ablation study of the skip-connection. The first row are content images, and we test the MDTN without skip-connection (the second row), without multiple style code (the third row) and with full structure (the forth row). It can be observed that the results without skip-connection cannot keep the content information from input sketches while the results multiple style code cannot maintain the style from the reference where obvious color differences can be found.}}
    \label{fig:structure_abl}
\end{figure}

Compared to general ways of decoding bottle neck feature code shown in Fig. \ref{fig:style_comparison}, MDTN combines multiple levels of both content and style information for the decoder, which can ensure the image generation based on sketches with different density levels. To further prove that this structure plays a significant role in the MDTN, we performed an ablation study. We first implemented the model without the skip connection, hence the content branch has to store all information in one latent code, which is then sent to the generator for the image reconstruction. Similarly, we also perform experiments without multiple style code, namely only using the bottle neck style code for decoding process. Results are shown in Fig.\ref{fig:structure_abl}. Without the skip connection, the results in different levels are almost identical and similar to the reference. While without the multiple style code, obvious color difference can be found between the generated images and the reference. This demonstrates that a single feature code is not capable to represent different levels of content information, especially in our case that sketch information are needed to be remained as much as possible. While with the skip connection and multiple style code, MDTN decoder selects and merges specific style features conditioned on content features in different levels, thus supports the multi-density translation. \par

\subsection{Why is the density encoder important in the MDSG?}
\begin{figure*}[ht]
    \centering
    \includegraphics[width=\textwidth]{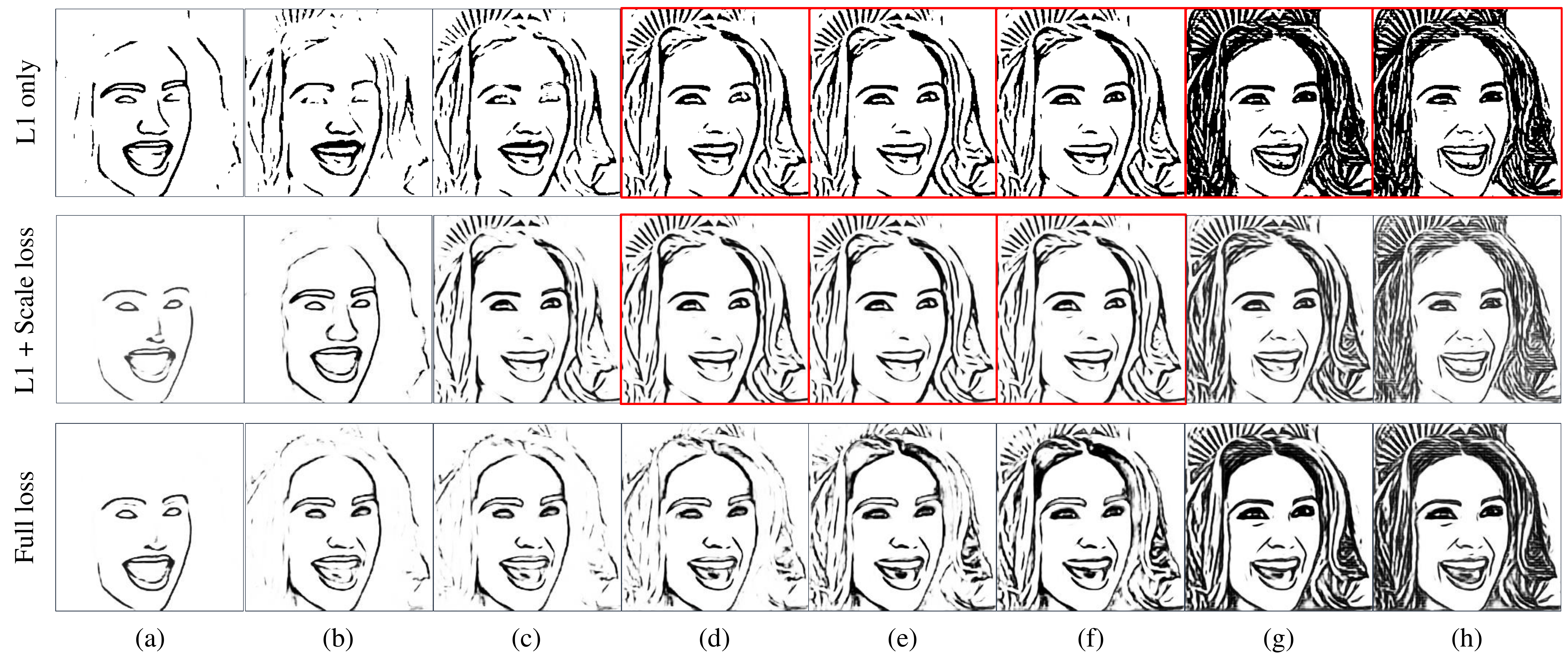}
    \caption{\small{We performed an ablation study to investigate the contribution of the density encoder, generated sketches (a)-(h) are based on scale from 0.1 to 0.8 with step 0.1, respectively. It shows that with the density encoder and its associated loss, the MDSG can achieve the best linearity. The red boxes indicate repeated images at different scales, which is lack of linearity.}}
    \label{fig:mscg_abl}
\end{figure*}

Ours MDSG can generate continuous sketches linearly related to the density factor and guide the continuous disentanglement in MDTN. This good property mainly comes from the density encoder which bridges the output image and the input density factor by imposing both density scale loss and AFD loss. To validate its effectiveness, we conducted an ablation study. We train the MDSG with three settings: 1) without the density encoder, 2) with the density scale loss defined by the density encoder 3) with both density scale loss and AFD loss defined by the density scale encoder. It can be observed from figure \ref{fig:mscg_abl}, if the MDSG is only constrained on some discrete density scales with L1 reconstruction loss, even though the continuous density scale mask is concatenated to the input image, it has a small impact on the latent space interpolation. The output changes suddenly from one discrete scale to another (1st row of figure \ref{fig:mscg_abl}). By introducing the density encoder, we can measure the scale reconstruction loss between the input scale and the decoded one. This loss helps to learn a shallow relation between the input density scale and the output sketch,  and thus enables some kinds of interpolation,  but there is no guarantee the interpolation is linear ($2^{nd}$ row of figure \ref{fig:mscg_abl}). This suggests us to add supervision in the feature space, forcing the latent space to be linear related to the density factor. The bottleneck feature in an auto-encoder network contains information about the complexity of the image, which can then be used for distance measurement. So we design the AFD loss to ensure the linearity between the bottleneck feature of generated sketch $G_c(M_{s'}, X)$ and that of its two neighbor key density sketch images $Y(s_i)$ and $Y(s_{i+1})$. It is shown our MDSG can achieve the highest linearity with both scale loss and AFD loss defined by the density encoder (3rd row of figure \ref{fig:mscg_abl}). 

\begin{figure}[ht]
    \centering
    \includegraphics[width=\columnwidth]{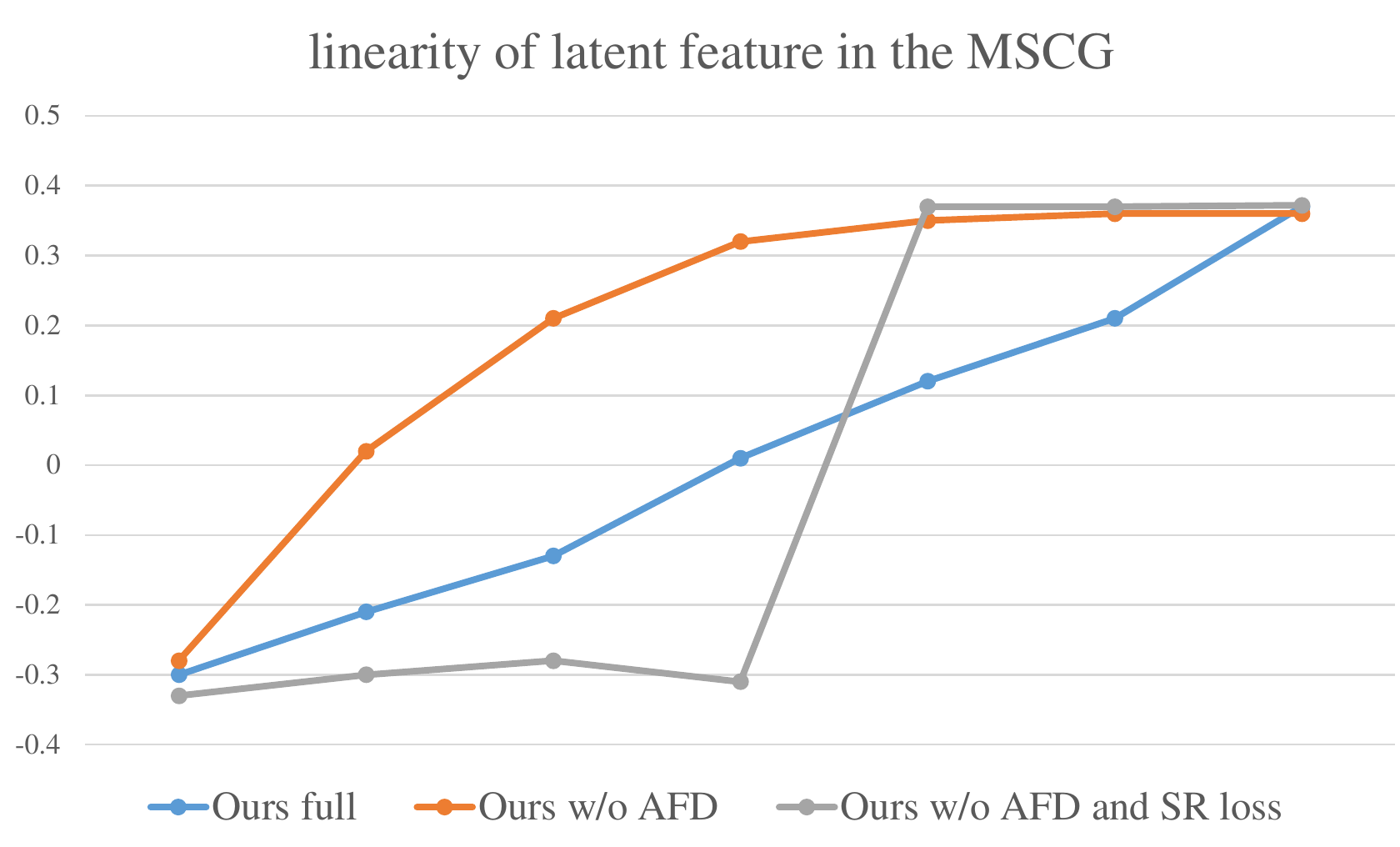}
    \caption{\small{To prove the linearity of the latent space in the MDSG, we extract the features from the encoder, use the PCA \cite{pca} to estimate the first principle component and plot it against the scale factors. The MDSG achieves the highest linearity with the full loss.}}
    \label{fig:linear_latent}
\end{figure}

To better visualize the linearity, we further extract the bottleneck features of $E_{G_c}$ with a list of scale factors from 0.2 to 0.8 where step $= 0.1$. And then, we apply PCA \cite{pca} to extract the first principle component and plot them against the corresponding density factors, shown in Fig.\ref{fig:linear_latent}. It also proves that our full model with both scale loss and AFD loss achieve the highest linearity. 


\subsection{Jointly training vs separately training}
We first train the MDTN and the MDSG separately, in this process, the MDTN is trained with only four density levels of sketches shown in fig.\ref{fig:face_content}. After that, we connect the  MDSG and the MDTN to fine tune the MDTN with continuously sampled random density factors. As we mentioned before, sketches generated by the MDSG are of geometric variation. Although such difference is quite small between two nearby density factors, they are still visually ineligible. With this joint training, the MDTN can access the content instances provided by the MDSG covering all the possible density scales in its defined domain. Therefore, the jointly fine-tuning helps the MDTN to disentangle an image continuously.\par

\begin{figure}[ht]
    \centering
    \includegraphics[width=\columnwidth]{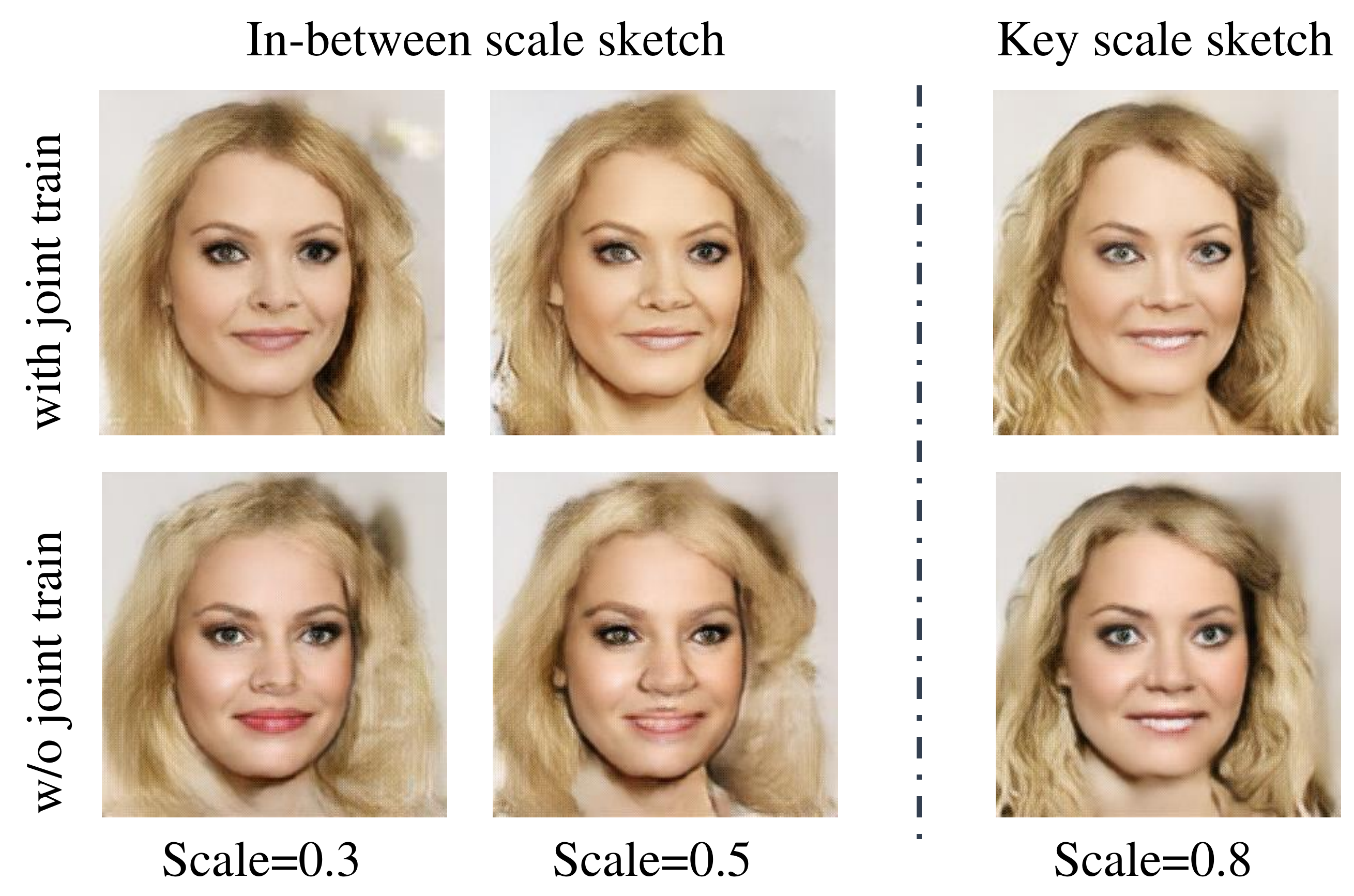}
    \caption{\small{Ablation study for joint training. For key density (e.g. 0.8), the quality of the generated images is similar with and without the joint-train, while at the in-between density (e.g. 0.3 or 0.5), the joint-training results are much better than separate-training results.}}
    \label{fig:joint_train_abl}
\end{figure}

To prove that the joint-train is an essential part, we show the results with and without joint-train in Figure \ref{fig:joint_train_abl}. We generate a set of sketches by the MDSG with the scale from 0.2 to 0.8, and test the MDTN with these sketches. We found that the key density results generated by the MDTN with and w/o joint-train are similar, while at the in-between density, results with the join-train have higher qualities. This is because the sketches in the in-between densities have their own distributions. As the model without fine-tuning has little knowledge about the certain distribution, it would be difficult to generalize to it. \par
\section{Experiment}
The implementation details, datasets, and other technical parameters in the training process are listed in the following section. And to evaluate our method including the capacity of multi-density S2I translation, quality and diversity of the generated results, etc, we design a group of experiments and compare with the state-of-the-art works qualitatively and quantitatively. 

\subsection{Dataset}
For the disentanglement of human face, we use the CelebA-HQ \cite{CelebAMask-HQ} dataset, which contains more than 200k celebrity images. To prove the robustness and generability of our model, we directly test both the MDSG and the MDTN without further finetuning on the FFHQ \cite{ffhq} dataset which consists of 700k images with higher diversity than the CelebA-HQ \cite{CelebAMask-HQ}. In the training and test process, we resize those images to 256x256. In addition to the human face datasets, we also train and test our model on the animation dataset Danbroo \cite{danbooru2018}, and similarly, we also test our model on the selfie2anime datasets \cite{selfie2anime}. For the Danbroo dataset, we use open source Mega face detection algorithm to detect and crop anime faces and adopt similar approaches mentioned previously to obtain four levels of sketches. Moreover, to prove that our method can also be applied to domains other than faces, we train our model on the SYNTHIA \cite{synthia} dataset, which can be further developed as a design tool.

\subsection{Implementation Details} 
Our framework contains two sub-network namely the MDSG and the MDTN. The MDSG contains a content generator and a density encoder, while the MDTN contains two encoders, a generator and a multi-scale discriminator similar to that used in the Pix2pixHD \cite{pix2pixhd}. We use the convolution blocks with ReLu as the activation layer and batch normalization for both sub-networks. The bottleneck features in the MDTN contains 512 channels, which are then concatenated with the features map extracted by the encoder at different levels. We use the batch size 4 and around 40 epochs and 20 epochs in training the MDTN and the MDSG, respectively. In the training of MDSG, we gradually increase the probability of middle scales. At the first epoch, we only train the key-frame sketches, and the middle scale is used incrementally in the later training process. With the pre-trained MDSG, it requires another 20 epoch for jointly training the MDTN in the joint fine-tuning stage. We trained our model on NVIDIA 2080Ti GPU, and it requires 4 days to pre-train the MDTN and 2 days for training the MDSG, and the fine-tuning stage needs another 4 days for the final results.\par 

\subsection{Baseline}
Many Image-to-image translation models can perform the sketch-to-image tasks. The prior work pix2pix \cite{pix2pix} can generate images based on a given sketch. BicycleGAN \cite{bicyclegan} proposed a way to generate multi-modal results based on sampled style codes. Later, DRIT \cite{drit} is proposed to achieve high quality multi-modal I2I translation with unpaired data. However, current state-of-the-art models are trained with sketch input in a single density level, to perform the comparison on sketches with different densities, we trained those models on the same training data for the MDTN. we compare our work with the aforementioned models and our framework achieves better results in both qualitative and quantitative aspects.

\subsection{Results}
\begin{figure*}[ht]
    \centering
    \includegraphics[width=\textwidth]{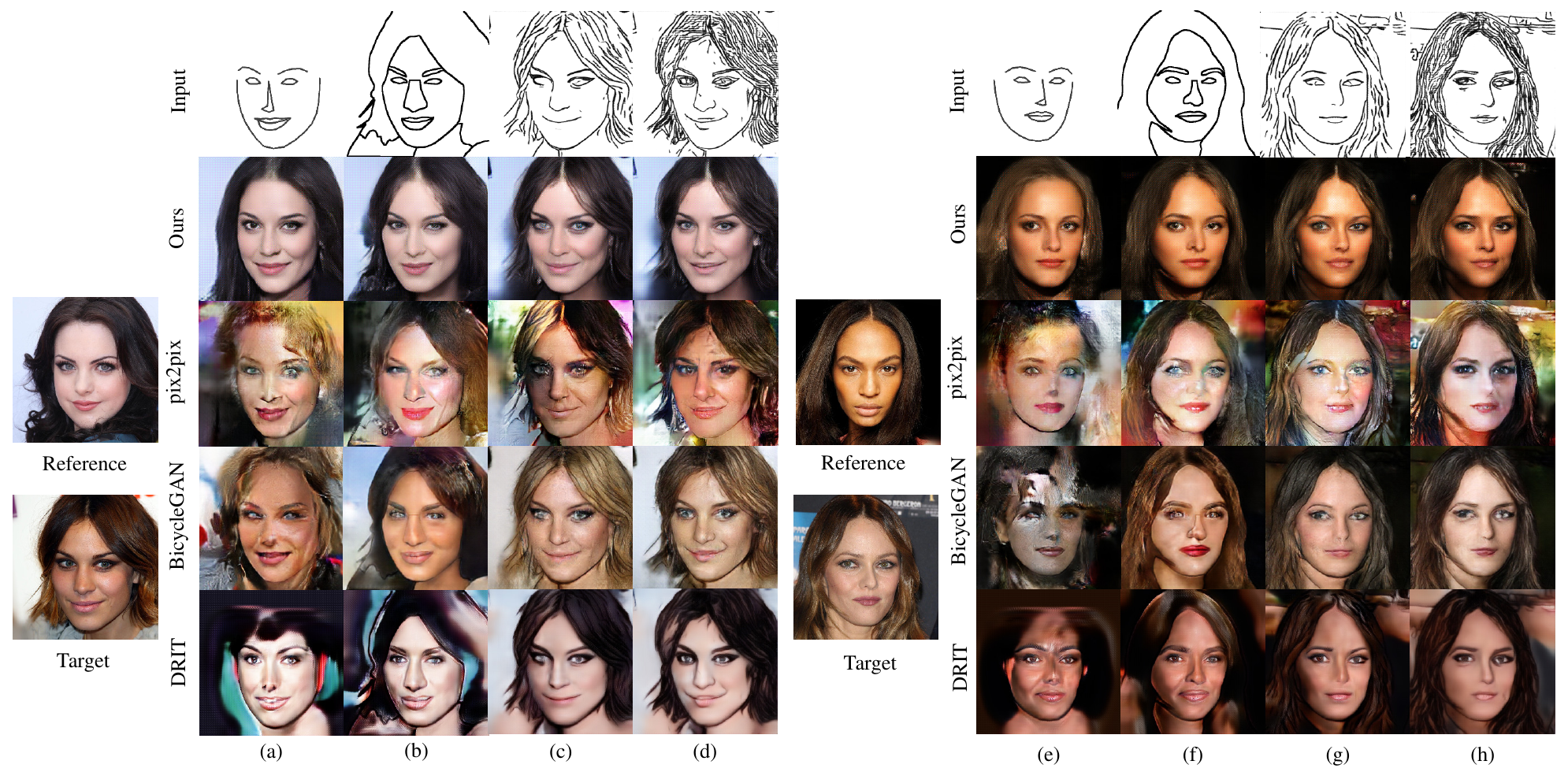}
    \caption{\small{Comparisons of our method with the baseline models in the sketch-to-image translation task. Four levels of sketches are used for training and testing. Our method demonstrates the best quality with multi-density sketches.}}
    \label{fig:compare}
\end{figure*}

 To compare with the baseline models, we train them with four key density sketches as our MDTN and then test them only with those key density sketches. The comparison results are shown in Figure \ref{fig:compare}. It is obvious that our method can better support multi-density S2I translation compared to other baseline models, pix2pix \cite{pix2pix} and BicycleGAN \cite{bicyclegan} cannot generate high quality images with low-density sketches. This is because both methods use a single level of feature code, as the information in the bottle neck feature code is limited, it is difficult to generate high quality output with a sparse input sketch. Moreover, these two methods adopt a skip-connection structure, which would take dominant control with a high-density sketch input. This can be observed in the left part (the forth row (d)) that the result is barely controlled by the given reference. As for the DRIT model which uses a single style code via AdaIn layers, though this ensures that the output can have certain similarities to the reference image, it cannot maintain some super-facial style details. Color and micro-stricture differences can be found between its results and the reference image.\par 
 
Besides the qualitative comparison, we also perform quantitative comparisons. Specifically, we randomly hold 800 pairs of CelebA-HQ and 4000 pairs of FFHQ, and test models on them with two modes. In the first mode, we use the same image as the content and style inputs to test the reconstruction capability, as shown in Table \ref{Table: Reconstruction}, while in the second mode, different images are used as the content and style inputs to test the disentanglement quality, as shown in Table \ref{Table: Disentangle}. Three metrics are adopted for the evaluation of the generated results. The structural similarity ($SSIM$) is used to compare the pixel-wise difference between the generated image and the ground truth. Perceptual image patch similarity ($LPIPS_{s}$) and Fréchet Inception Distance ($FID$) \cite{fid} are used to evaluate the reality of the results. Almost on every metric, our model could achieve the best performance in all the aspects of diversity and reality.\par  

\begin{figure*}[ht]
    \centering
    \includegraphics[width=\textwidth]{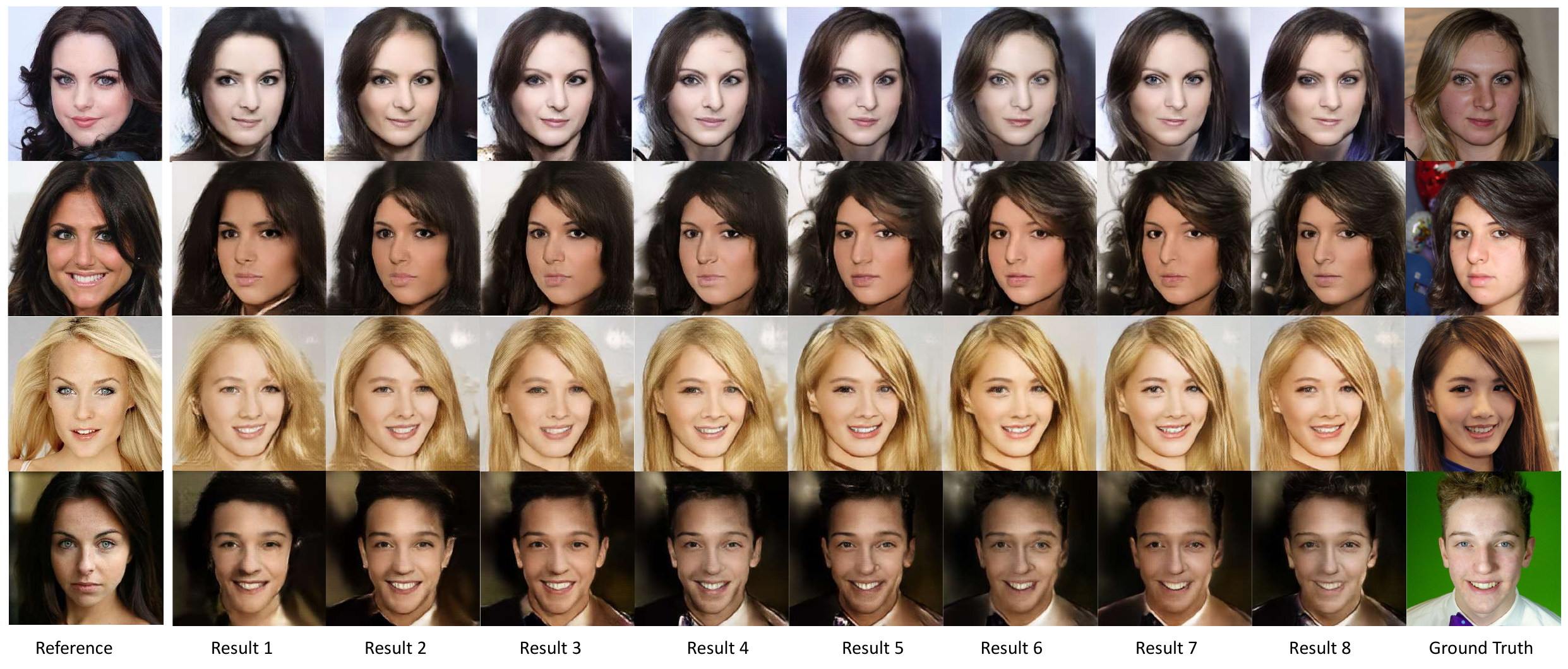}
    \caption{\small{Translation results based on sketches generated by MDSG. We use MDSG to generated different levels of sketches of the ground truth with a density factor from 0.1 to 0.8, which is similar to that in Fig.\ref{fig:mscg_abl} (third row). Those sketches are then sent to MDTN for the translation, results from left to right are generated based on low-density sketches to high-density sketches.}}
    \label{fig:joint_res}
\end{figure*}

\begin{figure*}[ht]
    \centering
    \includegraphics[width=\textwidth]{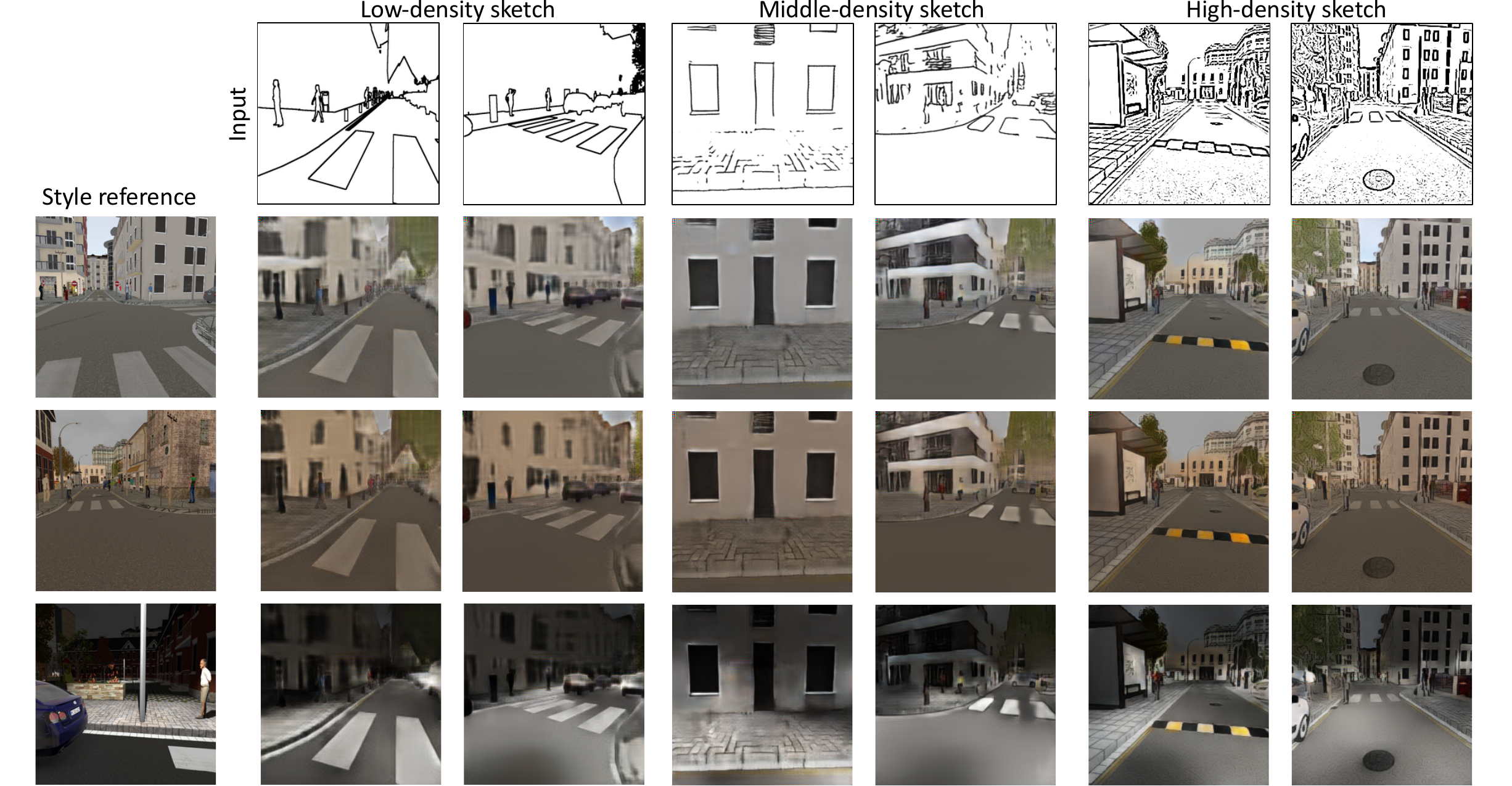}
    \caption{\small{Results of the SYNTHIA \cite{synthia} dataset. The  second and third columns show the results based on low-density sketches, and the next two columns are results generated based on middle-density sketches while the final two columns are results based on the high-density sketches.}}
    \label{fig:synthia}
\end{figure*}

In addition to support multiple key density sketches, out method can also perform S2I translation based on sketches with continuous densities generated from MDSG. We set the density factors from 0.1 to 0.8 with a step of 0.1, and use MDSG to generate sketches based those factors, the generated sketches are then fed into the jointly-train translation model to generate the final results, which are shown in Fig. \ref{fig:joint_res} Result1 to Result8. It can be observed that our model can successfully perform the S2I translation tasks with continuous density sketches. To further demonstrate the generality of our method, we also test the model on different datasets including the anime dataset Danbroo \cite{danbooru2018} and the SYNTHIA dataset \cite{synthia}, the multiple density S2I translation results are shown in Fig. \ref{fig:colorization} and Fig. \ref{fig:synthia}, respectively.

\begin{table*}[!ht]
\setlength{\tabcolsep}{0.8mm}{
\begin{tabular}{lllllll}
\hline
\multicolumn{1}{c|}{\multirow{2}{*}{Method}} & \multicolumn{3}{c|}{\textbf{CelebA-HQ}}                                                                        & \multicolumn{3}{c}{\textbf{FFHQ}}                                                                             \\
\multicolumn{1}{c|}{}                        & \multicolumn{1}{c}{$SSIM$ $\uparrow$}           & \multicolumn{1}{c}{$LPIPS_{s}$ $\downarrow$}          & \multicolumn{1}{c|}{$FID$ $\downarrow$}             & \multicolumn{1}{c}{$SSIM$ $\uparrow$}           & \multicolumn{1}{c}{$LPIPS_{s}$ $\downarrow$}          & \multicolumn{1}{c}{$FID$ $\downarrow$}             \\ \hline
\multicolumn{1}{c|}{\textbf{Pix2Pix}}        & \multicolumn{1}{c}{0.382}          & \multicolumn{1}{c}{0.456}          & \multicolumn{1}{c|}{109.349}         & \multicolumn{1}{c}{0.381}          & \multicolumn{1}{c}{0.498}          & \multicolumn{1}{c}{130.306}         \\
\multicolumn{1}{c|}{\textbf{BicycleGAN}}     & \multicolumn{1}{c}{0.485}          & \multicolumn{1}{c}{0.393}          & \multicolumn{1}{c|}{108.119}         & \multicolumn{1}{c}{0.491}          & \multicolumn{1}{c}{0.431}          & \multicolumn{1}{c}{120.721}         \\
%\multicolumn{1}{|c|}{\textbf{MUNIT}}     & \multicolumn{1}{c}{0.461}          & \multicolumn{1}{c}{0.433}          & \multicolumn{1}{c|}{105.172}         & \multicolumn{1}{c}{0.486}          & \multicolumn{1}{c}{0.443}          & \multicolumn{1}{c|}{120.811}         \\
\multicolumn{1}{c|}{\textbf{DRIT}}           & \multicolumn{1}{c}{0.465}          & \multicolumn{1}{c}{0.440}          & \multicolumn{1}{c|}{103.191}         & \multicolumn{1}{c}{0.477}          & \multicolumn{1}{c}{0.451}          & \multicolumn{1}{c}{121.127}         \\ 

\multicolumn{1}{c|}{\textbf{Ours w/o unet}}       & \multicolumn{1}{c}{0.541}          & \multicolumn{1}{c}{0.262} & \multicolumn{1}{c|}{56.102} & \multicolumn{1}{c}{0.553}          & \multicolumn{1}{c}{0.324} & \multicolumn{1}{c}{88.490} \\ 

\multicolumn{1}{c|}{\textbf{Ours w/o MS}}       & \multicolumn{1}{c}{0.472}          & \multicolumn{1}{c}{0.389} & \multicolumn{1}{c|}{71.310} & \multicolumn{1}{c}{0.516}          & \multicolumn{1}{c}{0.402} & \multicolumn{1}{c}{90.101} \\ \hline

\multicolumn{1}{c|}{\textbf{Ours}}       & \multicolumn{1}{c}{\textbf{0.551}}          & \multicolumn{1}{c}{\textbf{0.258}} & \multicolumn{1}{c|}{\textbf{56.264}} & \multicolumn{1}{c}{\textbf{0.555}}          & \multicolumn{1}{c}{\textbf{0.311}} & \multicolumn{1}{c}{\textbf{79.201}} \\ \hline
\end{tabular}}
\caption{\small{Reconstruction Results on CelebA-HQ and FFHQ testset. Bold denotes the best results, MS indicates multi-level style code used in the MDTN.}}
\label{Table: Reconstruction}
\end{table*}

\begin{table*}[!ht]
\setlength{\tabcolsep}{0.8mm}{
\begin{tabular}{c|ccc|ccc}
\hline
\multirow{2}{*}{Method} & \multicolumn{3}{c|}{\textbf{CelebA-HQ}} & \multicolumn{3}{c}{\textbf{FFHQ}} \\
                        & $LPIPS_{d}$ $\uparrow$   & $LPIPS_{s}$ $\downarrow$   & $FID$ $\downarrow$              & $LPIPS_{d}$ $\uparrow$  & $LPIPS_{s}$ $\downarrow$  & $FID$ $\downarrow$              \\ \hline
\textbf{BicycleGAN}     & 0.507    &    0.537      & 90.792            & 0.543  &   0.566     & 118.81          \\
%\textbf{MUNIT}     & 0.509    &    0.549      & 93.237            & 0.544  &   0.547     & 110.97          \\
\textbf{DRIT}           & 0.512    &     0.544     & 94.473            & 0.544  &   0.552     & 111.96           \\ 
\textbf{Ours w/o unet}     & 0.499    &    0.531      & 78.811            & 0.501  &   0.514     & 98.28          \\
\textbf{Ours w/o MS}     & 0.527    &    0.543      &   67.283          & 0.549  &   0.535     & 91.39          \\ \hline
\textbf{Ours}       & \textbf{0.530}    &   \textbf{0.432}       & \textbf{55.937}   & \textbf{0.550}  &   \textbf{0.470}     & \textbf{80.098}  \\ \hline
\end{tabular}}
\caption{\small{Sketch-to-Image translation results with references on CelebA-HQ and FFHQ testset. Bold denotes the best results, MS indicates multi-level style code used in the MDTN.}}
\label{Table: Disentangle}
\end{table*}
We also perform a quantitative evaluation for the baseline model and the proposed work to assess the generated results in terms of quality, diversity and reconstruction ability. We compare our model with conditional image synthesize methods Pix2Pix \cite{pix2pix} and BicycleGAN \cite{bicyclegan}, and DIRT \cite{drit}on CelebA-HQ and FFHQ datasets. Specifically, we randomly hold 800 pairs of CelebA-HQ and 4000 pairs of FFHQ for testing the appearance image reconstruction ability with corresponding content and the transfer generation performance while giving different appearance and content images.

Table \ref{Table: Reconstruction}. shows our quantitative comparisons with other baselines in term of image reconstruction ability. The structural similarity ($SSIM$) is used to compare the low-level difference between generated faces and original style faces. We also employ perceptual image patch similarity ($LPIPS_{s}$) and Fréchet Inception Distance ($FID$) \cite{fid} to evaluate the reality of the results. It is obvious that Pix2Pix obtain the worst performance under all metrics since no other extra guidance i.e. style reference could be used. By contrary, with the help of encoded style information, BicycleGAN and DIRT could have improved results. And the performance of our model without unet are less affected compared to that of removing the multi-level style code structure, since in the reconstruction experiments the reference and the input sketch have the same identity, therefore less effort is needed to ensure the output to be similar to both the input image and the reference. The leading scores demonstrated the superior ability of our model to yield images of good perceptual quality.

Besides the reconstruction results, we also show the the conditional image transferring performance in Table \ref{Table: Disentangle}. In the same manner we employ the $FID$ score to measure the distance between the feature distribution of generated faces and real face to evaluate the reality. $LPIPS_{d}$ is calculated to evaluate the diversity of generated results to prevent the mode collapse. What's more, to ensure the the consistency between generated results and corresponding style reference, we also calculate the $LPIPS_{s}$ to measure the similarity between them. As shown in Table. \ref{Table: Disentangle}, our model with the full structure could achieve the best performance in all the aspects of diversity, consistence and reality.

\section{Application}
In addition to general sketch-to-photo translation, we demonstrate some applications requiring multiple density sketch as input with our model.

\subsection{Multi-scale Face Editing}
\begin{figure}[ht]
    \centering
    \includegraphics[width=\columnwidth]{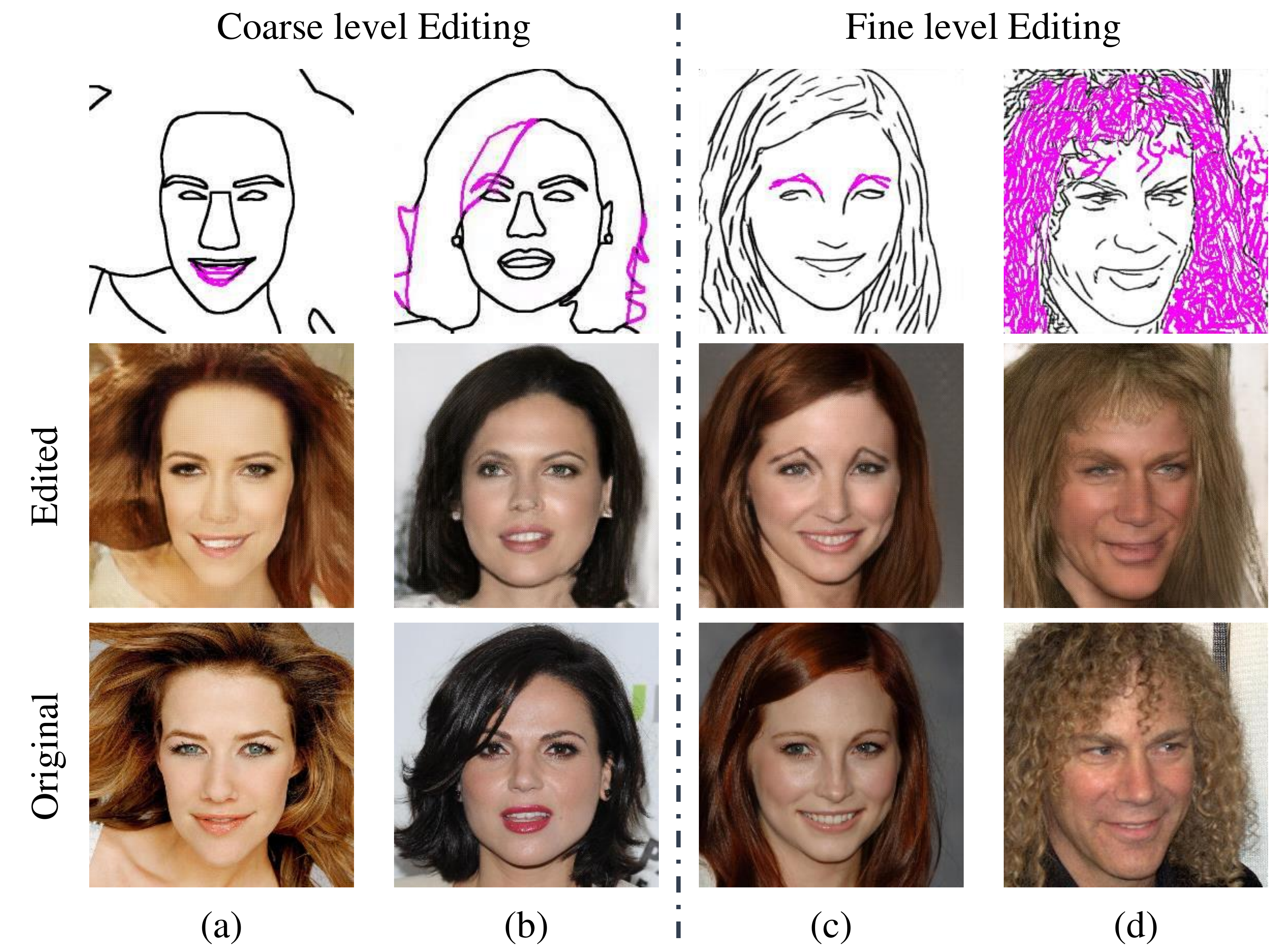}
    \caption{\small{Face Editing: our method supports face editing in multiple levels. In the coarse levels (a \& b), the user can edit outline of the face such as the shape of mouth and hair. In the fine levels (c \& d), the user can modify detailed textures.}}
    \label{fig:face_editing}
\end{figure}

The coarse level corresponding to a small $s_i$ allows the user to edit the large contours while ignoring the details which would be taken care of by the model. In the coarse level editing, the user can easily change the general characteristic of a human face, including face shape, length of hair, facial expression, etc., as shown in Fig. \ref{fig:face_editing} (a)(b). The fine Level corresponding to a large $s_i$ supports sophisticated manipulation on details such as the hair texture (curly to straight), and skin textures (adding or removing wrinkles), as shown in Fig. \ref{fig:face_editing} (c)(d). Compared to previous face editing work based on segmentation mask\cite{CelebAMask-HQ} or landmarks\cite{makeAface}, our method has two advantages. First sketching is a more intuitive and user-friendly way for image editing. Moreover, our method can control the editing process at different scales from major object boundaries to elaborated micro-structures.  

\subsection{Anime Colorization} 
\begin{figure}[ht]
    \centering
    \includegraphics[width=\textwidth]{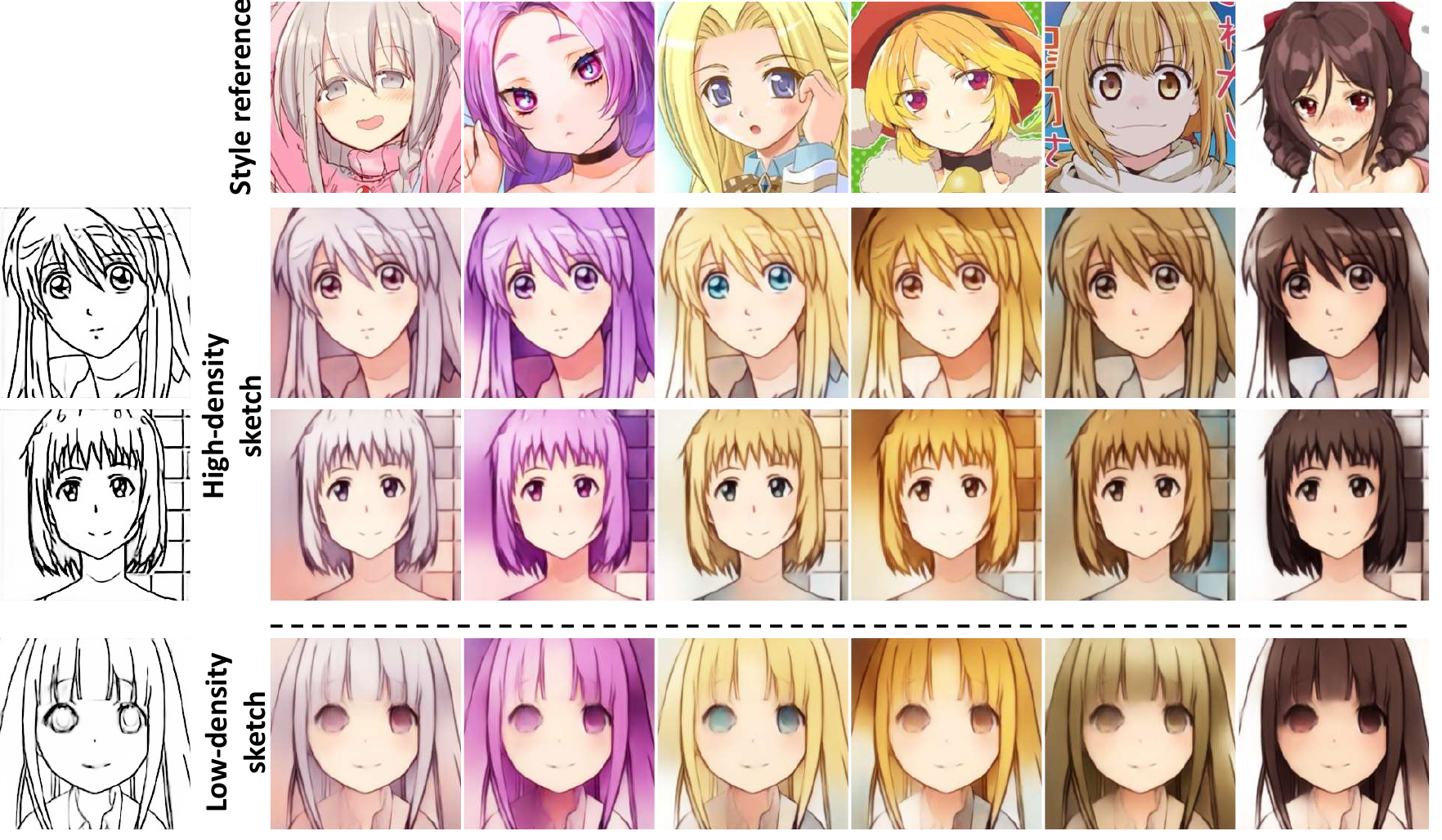}
    \caption{\small{Results of the anime colorization. The second and the third row represent the fine level colorization while the last row shows the coarse level colorization.}}
    \label{fig:colorization}
\end{figure}
Our method is general to different types of data, so it can be applied to Anime Colorization and Editing as well. Different from the previous sketch colorization method that our model can colorize images in both coarse and fine levels. In addition, we also support post-editing after colorization. Such editing allows not only rough modification but also detailed adjustments like adding shadows or highlights, modifying the minor textures, etc., results are shown in Fig.\ref{fig:colorization}. 

\section{Conclusion}
In this work, we have presented the first multi-density sketch-to-image translation framework. Moreover, we project the density scales into a continuous latent space, which can then be linearly controlled by a parameter. To achieve that, we designed a multi-density sketch generator which maps the control scale factors to the continuous content space; and a multi-density translation network that could generate images conditioned on different levels of sketches outputted by the MDSG. Our two sub-networks outperforms state-of-the-art methods on the human face dataset CelebA-HQ\cite{CelebAMask-HQ} and the FFHQ \cite{ffhq}. It is also proved that our method can be applied to other challenging datasets including Danbroo \cite{danbooru2018} and SYNTHIA \cite{synthia}. Finally, we demonstrate some applications including face editing and anime colorization, we supports those applications from coarse to fine levels. \par

% Bibliography
\bibliography{sample-bibliography}
\includegraphics[width=0.2\textwidth]{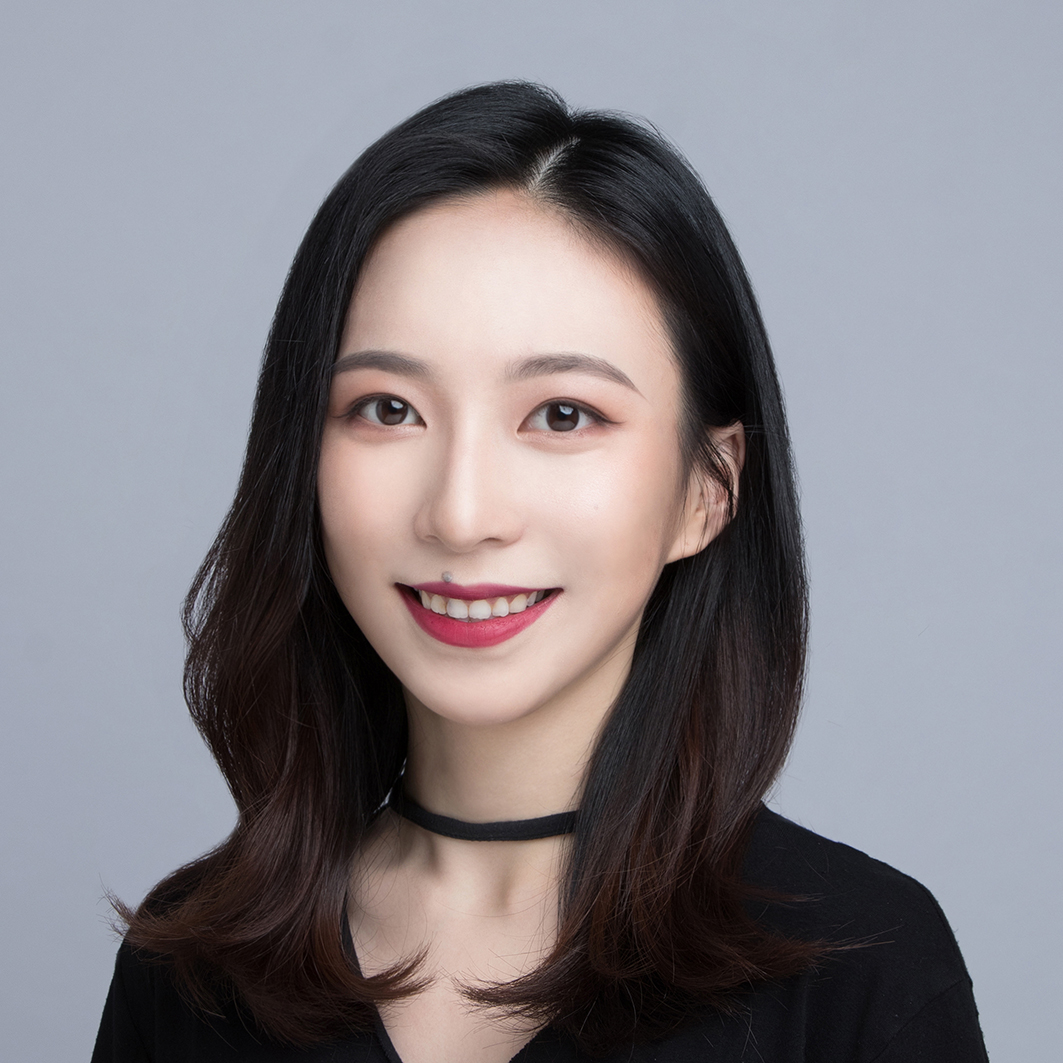}\small{\textbf{Jialu Huang} is a PhD student in the Department of Computer Science, City University of Hong Kong (CityU) since Sep 2018. She received the B.Eng. degree from Sun Yat-Sen University and MSc. degrees from King's College London. Her primary research interests fall in the fields of Image synthesis, Computer Graphics, Evolutionary Algorithm.}\par
\includegraphics[width=0.2\textwidth]{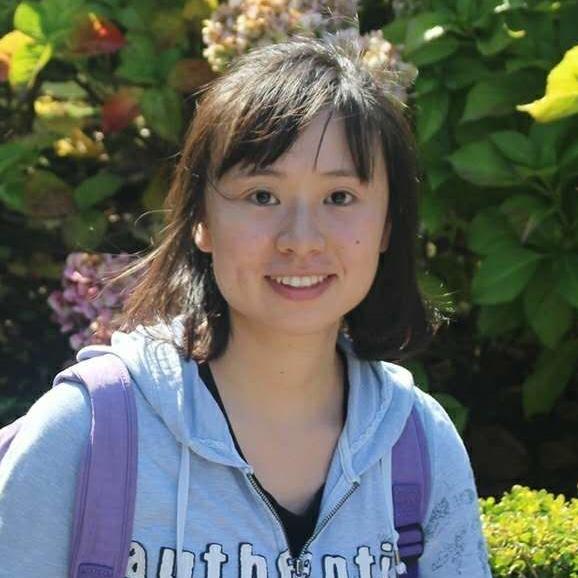}\small{\textbf{Liao Jing} is an Assistant Professor with the Department of Computer Science, City University of Hong Kong (CityU) since Sep 2018. Prior to that, she was a Researcher at Visual Computing Group, Microsoft Research Asia from 2015 to 2018. She received the B.Eng. degree from HuaZhong University of Science and Technology and dual Ph.D. degrees from Zhejiang University and Hong Kong UST. Her primary research interests fall in the fields of Computer Graphics, Computer Vision, Image/Video Processing, Digital Art and Computational Photography.}\par
\includegraphics[width=0.2\textwidth]{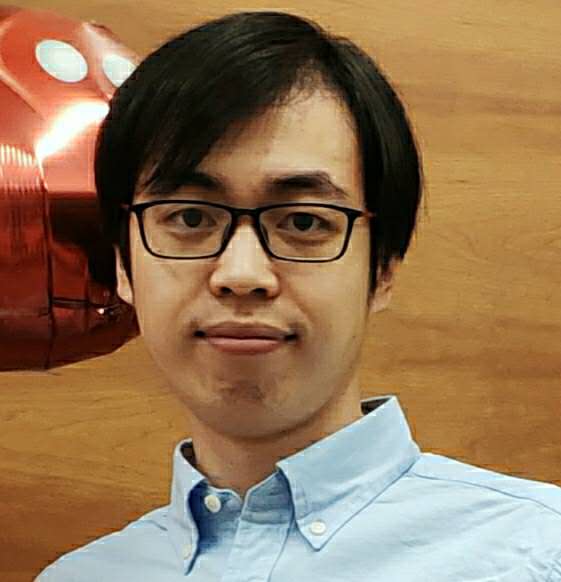}\small{\textbf{Tan Zhifeng} is an algorithm engineer of R\&D department of ASML corporation. He received the B.Eng. and MS. degree from Sun Yat-Sen University. His primary research interests fall in the fields of image processing and computer lithography. }\par
\includegraphics[width=0.2\textwidth]{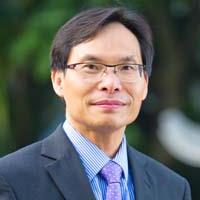}\small{\textbf{Sam Kwong} (Fellow, IEEE) received the B.Sc. degree in electrical engineering from the State University of New York, Buffalo, NY, USA, in 1983 and the M.Sc. degree in electrical engineering from the University of Waterloo, Waterloo, ON, Canada, in 1985, and the Ph.D. degree from the University of Hagen, Germany, in 1996. From 1985 to 1987, he was a Diagnostic Engineer with the Control Data Canada, Mississauga, ON, Canada. He later joined Bell Northern Research Canada, Ottawa, ON, Canada, as a Member of Scientific Staff, and the City University of Hong Kong (CityU), Hong Kong, as a Lecturer with the Department of Electronic Engineering in 1990. He is currently a Chair Professor with the Department of Computer Science, CityU. His research interests include  video coding, pattern recognition, and evolutionary algorithms. He is currently the Vice-President of Cybernetics with the IEEE SYSTEMS, MAN AND CYBERNETICS. He also serves as an Associate Editor of the IEEE TRANSACTIONS ON EVOLUTIONARY COMPUTATION, the IEEE TRANSACTIONS ON INDUSTRIAL ELECTRONICS, and the IEEE TRANSACTIONS ON INDUSTRIAL}\par

% Appendix
%\appendix
%\section{Appendix}

\end{document}